\ifdefined\XeTeXversion\else\pdfoutput=1\fi % arXiv: process with pdfLaTeX (no-op under XeTeX)
\documentclass{article}

\PassOptionsToPackage{numbers, compress}{natbib}

\usepackage[dblblindworkshop, final]{neurips_2026}
\workshoptitle{DiffuLM: Diffusion Language Models --- Foundations,
Efficiency, and Reasoning}

\usepackage[utf8]{inputenc} % allow utf-8 input
\usepackage[T1]{fontenc}    % use 8-bit T1 fonts
\usepackage{hyperref}       % hyperlinks
\usepackage{url}            % simple URL typesetting
\usepackage{booktabs}       % professional-quality tables
\usepackage{amsfonts}       % blackboard math symbols
\usepackage{nicefrac}       % compact symbols for 1/2, etc.
\usepackage{microtype}      % microtypography
\usepackage{xcolor}         % colors
\usepackage{amsmath,amssymb}
\usepackage{graphicx}

\newcommand{\cisep}{\textsc{ci-sep}}
\newcommand{\ns}{\textsc{n.s.}}
\newcommand{\base}{\textsc{base}}
\newcommand{\gma}{\textsc{m-dist}}   % integrated distilled checkpoint (G-M6)
\newcommand{\gmb}{\textsc{m-route}}  % routed RL, ratio 0 (G-M7)
\newcommand{\gmc}{\textsc{m-anchor}} % routed RL + anchor 0.1 (G-M8)
\newcommand{\gmd}{\textsc{m-comp}}   % domain-conditional composition (G-M9)

\title{Outcome-Routed Distillation and Reinforcement\\
Learning for Diffusion Language Models}

\author{%
  Wentao Lu\thanks{Equal contribution, Corresponding author} \\
  Celeris \\
  San Francisco, CA \\
  \texttt{toby@celeris.ai} \\
  \And
  Tianyu Zhu \\
  Celeris \\
  San Francisco, CA \\
  \texttt{alan@celeris.ai} \\
  \And
  Jesse Clark\thanks{Equal contribution} \\
  Celeris \\
  San Francisco, CA \\
  \texttt{jesse@celeris.ai} \\
}

\begin{document}
\maketitle

\begin{abstract}
On-policy distillation provides dense supervision on the student's own samples, but its targets still come from the teacher. Verifier reinforcement learning (RL) instead uses an outcome signal, but it provides no direct constraint on capabilities outside the verifier. Combining the two is difficult for diffusion language models (dLLMs), because bidirectional denoising makes the policy likelihoods required by ratio-based RL unreliable.

We develop a likelihood-free pipeline for a mixture-of-experts dLLM. Each rollout group is routed by verifier outcome. Mixed groups use ratio-free, group-centred REINFORCE. If every student rollout fails but a teacher succeeds, the model receives dense KL supervision on the student's denoising canvases. All other groups are skipped. A KL term anchored to the distilled checkpoint limits changes outside the verifier reward.

The tool-calling teacher succeeds on only $18.6\%$ of the training prompts. At 24 denoising steps, distillation exceeds the teacher by $6.77$ percentage points, and routing improves the distilled checkpoint by a further $18.31$ points. Routing lowers reasoning-trace accuracy by $0.71$ points but raises direct-answer accuracy on the same items by $13.21$ points. The anchored arm restores trace accuracy. Its tool-calling score differs by $-0.16$ points, with a confidence interval that includes zero, although this comparison also removes a small fallback distillation route. A domain-conditional recipe retains a $16.37$-point tool-calling gain; a second seed gives $18.23$ points and reproduces both preservation results, but the math result is seed-dependent. We pre-registered every arm and evaluated each comparison on paired items in a single sweep on one machine.
\end{abstract}

\section{Introduction}
\label{sec:intro}
On-policy distillation gives dense supervision on the student's own rollouts, but its targets remain teacher-derived~\citep{opdsurvey2026}. Verifier RL instead uses an outcome signal, but its sparse reward provides no direct constraint on capabilities outside the verifier. Recent work routes between these objectives for autoregressive models~\citep{srpo2026,dwyf2026,gac2026}. For dLLMs, bidirectional denoising makes the policy likelihoods required by PPO-style objectives unreliable~\citep{stabledrl2026,diffpo2025}. Objectives without importance-sampling ratios avoid this dependence~\citep{gdsd2026,wd12026}.

We develop a likelihood-free dLLM post-training pipeline with three components: multi-teacher on-policy distillation using dense KL on the student's denoising canvases, outcome-routed group-centred REINFORCE, and a KL anchor to the integrated distilled checkpoint (\gma). The anchor weight is set at each step from measured gradient norms. We fixed every criterion before the corresponding run and evaluated each comparison on paired items in a single-machine sweep. Throughout, \cisep{} denotes a 95\% confidence interval that excludes zero, and \ns{} denotes one that includes zero. Our contributions are:
\begin{itemize}
    \item A \emph{likelihood-free training pipeline} for dLLMs that combines multi-teacher on-policy distillation, outcome-routed REINFORCE, and measured KL anchoring without policy-likelihood ratios.
    \item An \emph{empirical routing criterion}: routing helps when the teacher can solve few of the student's remaining failures and hurts when it replaces a teacher that can still provide successful targets.
    \item A \emph{capability-preservation analysis}. Routed RL lowers reasoning-trace accuracy while raising direct-answer accuracy on the same items. The anchored arm restores trace accuracy without a CI-separated tool-calling loss.
    \item A \emph{domain-conditional composition} based on this criterion. Two seeds reproduce the tool-calling gain and both preservation results. Math varies by $2.1$--$2.5$ points across seeds, and one seed falls below \gma{} in one cell.
    \item A \emph{measurement of anchor strength}. Maintaining a $0.1$ gradient-norm ratio requires a loss coefficient that varies by $216\times$ within one run. The between-arm anchor effects remain provisional because \gmc{} also removes a small fallback distillation route.
    \item A \emph{pre-registered evaluation protocol} with paired per-item tests, measured training variation, and instrument-validity checks (\S\ref{sec:discipline}).
\end{itemize}

\section{Related work}
\label{sec:related}

\textbf{RL--distillation routing.} SRPO~\citep{srpo2026} routes correct samples to GRPO and failures to distillation. Distill-Where-You-Fail~\citep{dwyf2026} (concurrent) adds teacher guidance to all-fail RL groups, and GAC~\citep{gac2026} adaptively mixes SFT and RL. These methods target autoregressive models and do not anchor to an integrated multi-teacher checkpoint.

\textbf{On-policy distillation.} Prior work identifies teacher-boundedness as a central limitation of on-policy distillation~\citep{opdsurvey2026,opdfailures2026}. MOPD~\citep{mopd2026} provides our multi-teacher basis. RAFT~\citep{raft2026} (concurrent) conditions the teacher on the gold answer over top-$K$ targets.

\textbf{RL for dLLMs.} Existing work focuses on estimating likelihood ratios under diffusion sampling~\citep{stabledrl2026,tracerl2026,diffpo2025}, although objectives without importance-sampling ratios have also been proposed~\citep{gdsd2026,wd12026}. Our pipeline uses no policy-likelihood ratios.

\textbf{Capability retention.} KL regularisation to a reference model is standard in RLHF. CoKL~\citep{cokl2026} (concurrent) shows that full-distribution KL can also preserve the reference model's errors, a possible mechanism for our step-dependent result (\S\ref{sec:analysis}). Prior work intervenes at the token~\citep{talr2025}, example~\citep{ogs2026}, mixture~\citep{dynamics2026}, and architecture~\citep{memsft2026} levels. Our anchor operates at the objective level and sets its weight from a measured per-step gradient-norm ratio, extending GradNorm-style balancing~\citep{gradnorm2018}.

\section{Setting and method}
\label{sec:method}

\textbf{Setting.} We post-train DiffusionGemma-26B-A4B~\citep{diffusiongemma2026}, a mixture-of-experts dLLM with 26B total and 4B active parameters, for 240 optimiser steps per arm. We use three domains: math ($608$ held-out items, exact match), workflow ($1{,}399$ items, continuous score), and tool-calling ($1{,}240$ items, function-call match). Each arm is evaluated at 8, 24, and 48 denoising steps, with 24 as the operating point. Every arm uses AdamW at learning rate $10^{-5}$ on a cosine schedule with five warmup steps, gradient-norm clip $20$, eight rollout groups of $k{=}8$ per optimiser step, and a 2048-token sequence length with a 1792-token content budget. To measure capability retention, we evaluate MMLU-Pro~\citep{mmlupro2024} on the same $12{,}032$ items with and without chain-of-thought. We also score a fixed set of 200 multi-turn tool-execution items because the training prompts are single-turn.

\textbf{Domains and data.} Math uses GSM8K~\citep{gsm8k2021} and MATH~\citep{mathdataset2021}, with boxed chain-of-thought answers; the evaluation set contains 608 held-out items. Workflow combines INTERS~\citep{inters2024}, Super-NaturalInstructions~\citep{superni2022}, UniversalNER~\citep{universalner2023}, and P3~\citep{promptsource2022}. We report the mean of each dataset's primary metric. This score is continuous, so we make no paired significance claims. Tool-calling uses a single-turn function-calling training pool from Nemotron (CC-BY-4.0) and BFCL v3 AST match for evaluation; BFCL is not used in training. We remove training rows that share an 8-gram with the evaluation set or exceed the sequence budget.

\begin{figure}[t]
\centering
\includegraphics[width=\linewidth]{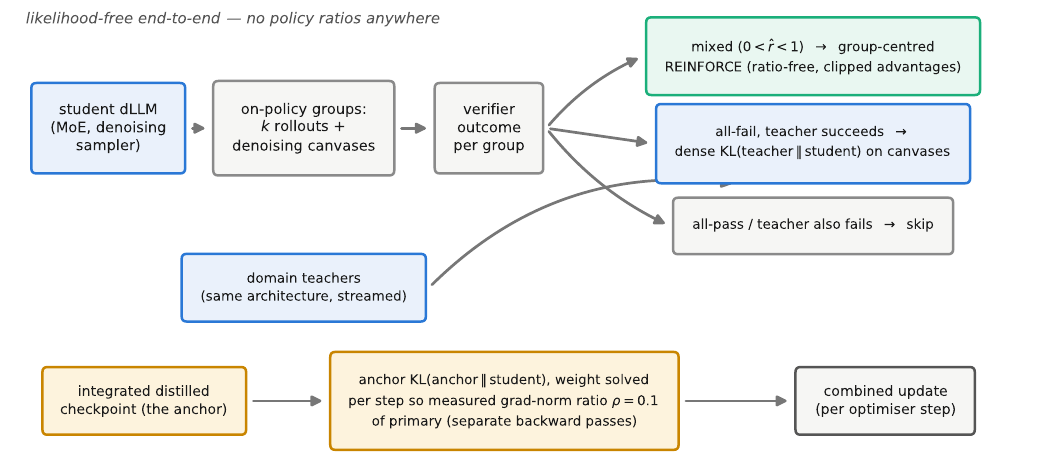}
\caption{Likelihood-free training pipeline. Mixed groups use ratio-free,
group-centred REINFORCE. All-fail groups use dense canvas KL only when a
teacher succeeds. All-pass groups and all-fail groups without a successful
teacher are skipped. The KL anchor to \gma{} is scaled each step to a measured
$0.1$ gradient-norm ratio using separate backward passes.}
\label{fig:overview}
\end{figure}

\begin{table}[t]
\centering
\small
\caption{Training arms. Each row changes one component relative to the row above.}
\label{tab:arms}
\setlength{\tabcolsep}{4pt}%
\begin{tabular}{@{}p{1.9cm}p{7.3cm}p{3.5cm}@{}}
\toprule
arm & objective & change from previous \\
\midrule
\gma & dense multi-teacher KL on the student's denoising canvases, with one same-architecture teacher per domain & baseline \\
\gmb & route by verifier outcome: mixed groups use ratio-free REINFORCE; all-fail groups use dense KL when a teacher succeeds; all other groups are skipped & $+$ routing \\
\gmc & add KL to \gma{} at a measured $0.1$ gradient-norm ratio; math and tool-calling teachers are not loaded, so the fallback route is unavailable & $+$ anchor, $-$ fallback route \\
\gmd & math uses dense KL to \gma{} as its primary objective; tool-calling and workflow follow \gmc{} & domain-conditional objective \\
\bottomrule
\end{tabular}
\end{table}

\textbf{\gma{} --- multi-teacher on-policy distillation.} \gma{} applies dense position-wise $\mathrm{KL}(p_T\|p_\theta)$ to the student's own denoising canvases, with one same-architecture teacher per domain, following MOPD~\citep{mopd2026}. The objective requires no policy-likelihood ratios. We trained all teachers in-house from the same base model during earlier single-domain phases. Each teacher uses the method that performed best in its domain: verifier RL for math ($+6.9$ over \base, \cisep) and tool-calling ($+8.6$, \cisep), and SFT for workflow ($+3.8$). No external model is used (Table~\ref{tab:teachers}). The tool-calling teacher improves the benchmark but succeeds on only $18.6\%$ of its training pool. It still provides dense token-level targets, but successful completions are rare.

\begin{table}[t]
\centering
\small
\caption{Teacher provenance. Each teacher starts from \base{} and is
post-trained in an earlier single-domain experiment using the best method for
that domain. Mixed groups satisfy $0<\hat r<1$. Each RL teacher consumes every
mixed group once. Benchmark gains and inheritance ratios use the
same distillation-era scoring sweep; Table~\ref{tab:gate} reports a later
sweep. No table combines scores from different sweeps.}
\label{tab:teachers}
\setlength{\tabcolsep}{4pt}%
\begin{tabular}{@{}p{1.4cm}p{3.4cm}p{4.3cm}rp{2.2cm}@{}}
\toprule
teacher & method & training data & steps & benchmark gain vs \base \\
\midrule
math & verifier RL: per-position GDSD variant~\citep{gdsd2026}, frozen experts, init from base &
$\sim$200--320 competition-math prompts consumed of 5{,}833 sampled
(GSM8K~\citep{gsm8k2021}+MATH~\citep{mathdataset2021}), $G{=}16$, mixed-outcome groups only & 25--40 &
$+6.9$ @24 (\cisep{} at 8/24/48) \\
\addlinespace
workflow & SFT: frozen-experts full fine-tune, seq-2048, thought-channel
format & 30.5M tokens; native long-CoT + pass@$k$-harvested completions & 199 &
$+3.8$ macro (classification $+12$) \\
\addlinespace
tool-calling & verifier RL: the same GDSD variant, frozen experts, init from base;
strict+graded AST-multiset verifier & 350 single-turn prompts
$\to$ 232 mixed groups ($G{=}8$, mixed denoising steps $\{8,24,48\}$), each
visited once & 29 & $+8.6$ @24 (\cisep{} at 8/24/48) \\
\bottomrule
\end{tabular}
\end{table}
Using the distillation-era sweep, the inheritance ratios (student gain divided by teacher gain) are $0.90$ for math, $0.82$ for workflow, and $1.92$ for tool-calling. The first two remain below the teacher's gain. For tool-calling, the student's gain over \base{} is larger than the teacher's, so the ratio exceeds one.

For the later comparison, we re-scored all teachers in the same sweep as the students. At 24 steps, \gma{} scores $70.24$ on tool-calling, compared with $63.47$ for its teacher, a gain of $6.77$ points. The difference is CI-separated at all three step counts ($+7.18 / +6.77 / +6.69$), and the in-sweep inheritance ratio is $1.68$. The re-scored math teacher scores $71.05$ at 24 steps and does not separate from either \gma{} or \gmd{}. On workflow, \gmd{} and its teacher differ by less than $0.005$ at every step count. We report no in-sweep inheritance ratio for math or workflow because neither student--teacher difference is CI-separated.

\textbf{\gmb{} --- outcome-conditional routing.} Each on-policy group contains $k{=}8$ rollouts and is routed by verifier outcome (Figure~\ref{fig:overview}). Mixed groups ($0<\hat r<1$) use ratio-free group-centred REINFORCE. If all student rollouts fail but a teacher succeeds, the group uses dense KL. All-pass groups and all-fail groups without a successful teacher are skipped. The RL term is
\begin{equation}
\mathcal{L}_{\mathrm{RL}} = -\frac{1}{D}\sum_{j} A_j \sum_{i \in S_j} \log p_\theta(y_i),
\qquad A_j = \mathrm{clip}\!\left(r_j - \hat r,\, -1,\, 1\right),
\label{eq:rl}
\end{equation}
where $j$ indexes the group's rollouts, $S_j$ is rollout $j$'s supervised canvas positions, and $D$ is the step's total supervised-position count. $A_j$ is detached and computed from the full group before eligibility filtering; the clip does not bind for binary rewards. The objective uses only $\log p_\theta$ on realised tokens; it contains no old-policy forward pass and no $\exp(\log p_\theta - \log p_{\text{old}})$. A preliminary measurement found that BF16 diffusion replay drifts by $\sim0.1$ nat, making $p_{\text{old}}$ unstable, so ratio-based objectives are excluded.

\textbf{\gmc{} --- measured anchoring.} \gmc{} adds a full-vocabulary $\mathrm{KL}(p_{\gma}\|p_\theta)$ term on the training-batch canvases. At each step, its weight is set so that the measured anchor gradient norm is $\rho=0.1$ of the primary-objective norm. This requires a separate backward pass for each objective. We record both norms and their cosine at every step. The realised ratio is $0.1000$ for all 240 steps.

\gmc{} also differs from \gmb{} in teacher availability. To fit the anchor in memory, we do not load the math or tool-calling teachers, so the fallback distillation route is unavailable. That route fires for $0.6\%$ of tool-calling groups and $1.8\%$ of math groups in \gmb{}. The comparison therefore combines the addition of the anchor with removal of the fallback route, and the data do not isolate their effects. We treat the anchor-scale differences in Table~\ref{tab:tradeoff} as provisional. \gmd{} uses the same teacher-loading configuration as \gmc{}, so their comparison does not contain this confound.

\textbf{\gmd{} --- domain-conditional composition.} \gmd{} changes only the math objective relative to \gmc{}. Math uses dense $\mathrm{KL}(p_{\gma}\|p_\theta)$ as its primary objective, making \gma{} its teacher. Tool-calling and workflow are unchanged. This choice follows the route proportions in \S\ref{sec:results}: math uses majority RL and falls below \gma{} in both routed arms at 24 and 48 steps, whereas workflow uses only distillation and does not regress. The composition targets the 24- and 48-step math deficits but gives up the routed arms' 8-step advantage. Thus \gma{} serves as the math teacher and as the $\rho{=}0.1$ anchor for tool-calling.

\section{Experimental discipline}
\label{sec:discipline}

Before each run, we committed a pre-registration covering acceptance criteria, intervention levels, fixed endpoints, and stopping rules. Comparisons use paired per-item tests: exact McNemar's test for accuracies and a bootstrap for continuous scores, both with 95\% confidence intervals. We withhold a cell if more than 5\% of its outputs are truncated or fewer than 90\% parse. Checkpoints that fail a preservation criterion remain in the analysis. All arms within a comparison are trained and scored on one GH200-class GPU. Each training arm runs for 240 optimiser steps ($\approx1$--1.5 days), and each scoring sweep takes $\approx14$\,h.

\textbf{Evaluation and training noise.} Paired per-item tests control evaluation noise but not variation from training. A previous three-run distillation experiment, using different prompt slices, varied by $0.16$ points on the same 608-item math set. To measure variation for a routed recipe, we trained \gmd{} a second time with identical prompt slices and changed only the run seed. Both seeds were scored in one four-arm sweep (\S\ref{sec:results}). The seed-to-seed spread is $0.14$--$0.19$ points on the preservation evaluations, $1.6$--$1.9$ on tool-calling, and $2.14$--$2.47$ on math. Math variation for the routed recipe is therefore $13$--$15\times$ the distillation reference. The $0.16$-point tool-calling difference between \gmb{} and \gmc{} is much smaller than the $1.6$--$1.9$-point tool-calling spread between the two \gmd{} seeds. Because that spread comes from a different recipe, it provides only an order-of-magnitude reference.

Cross-sweep comparisons are also unreliable. The same base model scores $65.62$--$67.76$ at 24 steps across machines, a $2.14$-point shift that exceeds the anchor-scale effects ($0.16$ and $0.98$). A paired same-machine comparison cancels a shift common to both arms. We therefore report only differences measured within one sweep on one machine. The full reproducibility record will be released upon publication.

\section{Results}
\label{sec:results}

All scores range from $0$ to $100$. Math is pass@1 exact match. Tool-calling is BFCL v3 AST match. The two MMLU-Pro preservation evaluations and the multi-turn set report accuracy, all in percent. Workflow is the mean of the four datasets' primary metrics, scaled by $100$; it is not a success rate.

\begin{table}[t]
\centering
\small
\caption{Scores on the three training domains, measured in one paired,
single-machine sweep. Math is pass@1 accuracy; workflow is a continuous
macro score scaled by $100$; tool-calling is AST-match accuracy. Bold marks
\gmd{}. For seed 1, none of its paired accuracy cells is CI-separated below
\gma{}, and all workflow point estimates are at least as high.}
\label{tab:gate}
\begin{tabular}{llrrrrr}
\toprule
domain & steps & \base & \gma & \gmb & \gmc & \gmd \\
\midrule
math          & 8  & 57.89 & 60.20 & 62.34 & 62.34 & \textbf{58.72} \\
              & 24 & 65.62 & 72.53 & 69.74 & 70.07 & \textbf{70.39} \\
              & 48 & 66.28 & 73.36 & 70.72 & 69.90 & \textbf{71.71} \\
workflow      & 8  & 43.98 & 49.88 & 50.60 & 51.12 & \textbf{51.48} \\
              & 24 & 45.47 & 50.33 & 51.09 & 51.60 & \textbf{51.60} \\
              & 48 & 45.28 & 49.61 & 50.97 & 50.79 & \textbf{51.02} \\
tool-calling  & 8  & 51.85 & 67.50 & 86.45 & 85.32 & \textbf{85.16} \\
              & 24 & 53.55 & 70.24 & 88.55 & 88.39 & \textbf{86.61} \\
              & 48 & 55.81 & 70.32 & 89.11 & 89.03 & \textbf{87.50} \\
\bottomrule
\end{tabular}
\end{table}

\begin{table}[t]
\centering
\small
\caption{Key differences from \gma{} for \gmb{} (anchor ratio $0$) and
\gmc{} (ratio $0.1$). Deltas use full-precision paired scores and can differ
in the last digit from the rounded values in Table~\ref{tab:gate}. \gmc{}
also lacks the fallback distillation route available to \gmb{}, so the
between-arm differences do not isolate the anchor.}
\label{tab:tradeoff}
\begin{tabular}{lll}
\toprule
cell (difference from \gma) & ratio $0$ & ratio $0.1$ \\
\midrule
tool-calling @24 & $+18.31$ \cisep & $+18.15$ \cisep{} (difference from ratio 0: $-0.16$, \ns) \\
math @8          & $+2.14$ \ns{}   & $+2.14$ \ns \\
math @24         & $-2.80$ \cisep  & $-2.47$ \ns{} (difference: $+0.33$) \\
math @48         & $-2.63$ \cisep  & $-3.45$ \cisep{} (difference: $-0.82$) \\
preserve CoT vs.\ \base & $-0.71$ \cisep & $+0.27$ \ns{} (difference: $+0.98$, \cisep) \\
multi-turn set vs.\ \base & $+7.00$ \cisep & $+9.50$ \cisep \\
\bottomrule
\end{tabular}
\end{table}

\begin{figure}[t]
\centering
\includegraphics[width=\linewidth]{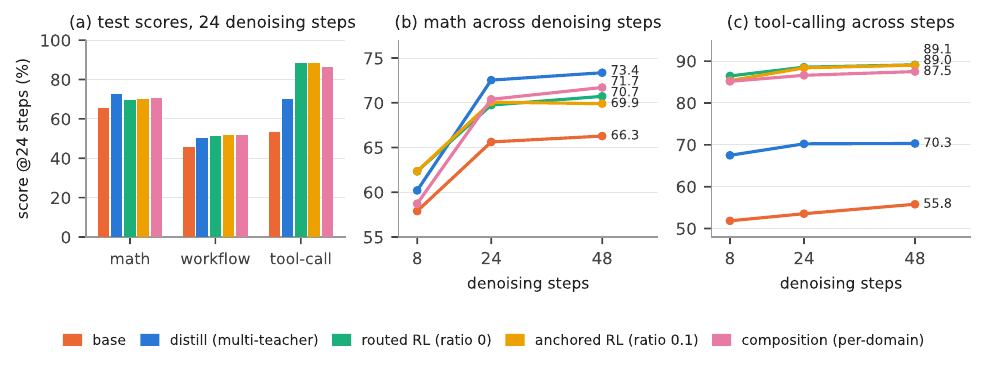}
\caption{(a) Scores at 24 denoising steps. (b, c) Scores across step counts.
Tool-calling gains persist for every routed arm. Math distinguishes the
methods: \gmc{} narrows the deficit at 24 steps and widens it at 48, while
\gmd{} remains close to \gma{} at 24 and 48 steps but gives up the routed
arms' $2.14$-point advantage at 8 steps, ending $1.48$ points below \gma{}
(\ns).}
\label{fig:results}
\end{figure}

\textbf{Routing improves tool-calling beyond distillation and the teacher.} The tool-calling teacher succeeds on $18.6\%$ of the training prompts, so most on-policy groups have no successful teacher completion. \gma{} still applies dense KL to every non-prompt position and scores $70.24$. \gmb{} uses verifier feedback on mixed groups and scores $88.55$, a gain of $18.31$ points over \gma{} (CI $[15.70,20.91]$, $n{=}1240$) and $25.08$ points over the re-scored teacher ($63.47$; \S\ref{sec:method}). This gain is accompanied by math decreases of $2.80$ points at 24 steps and $2.63$ points at 48 steps, both CI-separated. At 8 steps, math instead increases by $2.14$ points (\ns). \gmc{} retains an $18.15$-point tool-calling gain (CI $[15.62,20.67]$) and restores reasoning-trace accuracy (below), but math remains below \gma{} at 48 steps (Table~\ref{tab:tradeoff}; Figure~\ref{fig:results}).

\textbf{Teacher coverage tracks the effect of routing.} The routing rule is fixed; the observed proportions are determined by student and teacher outcomes. For tool-calling in \gmb{}, $10.0\%$ of groups are mixed and use RL, $18.2\%$ are all-pass, $71.2\%$ are all-fail with a failed teacher and are skipped, and $0.6\%$ are all-fail with a successful teacher and use KL. Among groups that the student fails completely, the teacher supplies a successful completion for only $0.8\%$. Math has the opposite profile: $54.6$--$61.1\%$ of groups use RL, the teacher usually succeeds, and both routed arms fall below \gma{} at 24 and 48 steps. At 8 steps, neither routed arm is below \gma{}. Workflow uses distillation for every group and holds or improves in every arm. Across domains, teacher coverage of the student's failures tracks the effect of routing; the fraction of groups assigned to RL does not.

\begin{figure}[ht]
\centering
\includegraphics[width=0.92\linewidth]{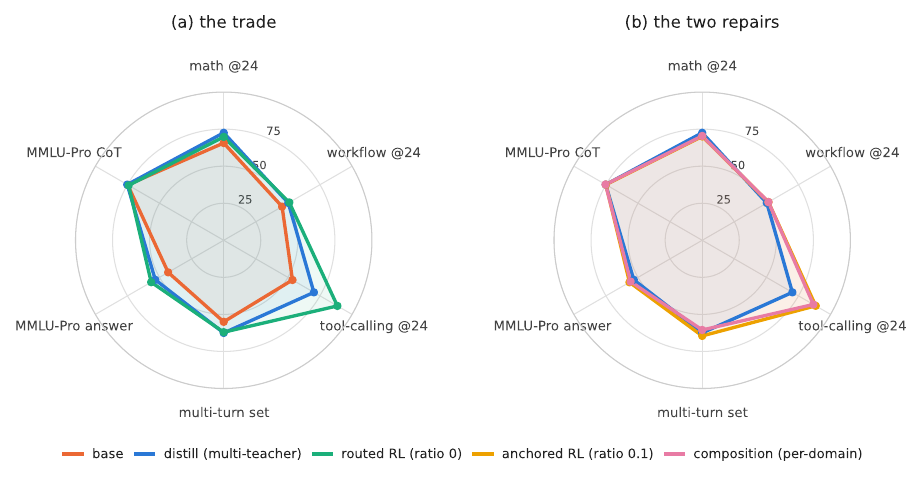}
\caption{Capability scores on a common 0--100 scale. (a) Relative to \base{}
and \gma{}, routed RL improves tool-calling while reducing math and
chain-of-thought accuracy. (b) The anchored and composed arms retain most of
the tool-calling gain while recovering part of the math and chain-of-thought
losses.
Each panel contains three polygons to keep every pair distinguishable under
simulated colour-vision deficiency. Table~\ref{tab:gate} and
Figure~\ref{fig:results} show all five arms.}
\label{fig:radar}
\end{figure}

\textbf{Routing changes reasoning traces without reducing answer accuracy.} Relative to \base{}, \gmb{} lowers free-form chain-of-thought accuracy by $0.71$ points ($p{=}.016$, $n{=}12{,}032$ paired) but raises direct-answer accuracy on the same items by $13.21$ points (Figure~\ref{fig:radar}). Multi-turn accuracy is $62.0$, above \base{} ($55.0$) and not CI-separated below \gma{} ($62.5$). The verifier scores final answers but not reasoning traces, so this pattern is consistent with a narrow change to unscored traces rather than a broad loss of answer capability. \gmc{} raises trace accuracy by $0.98$ points relative to \gmb{} (\cisep, $p{=}.0005$) while retaining the direct-answer gain ($+13.36$ points). Its multi-turn score is $64.5$, the highest of the four trained arms.

\textbf{Domain-conditional composition reduces the math deficit.} At 24 and 48 steps, \gmd{} scores $2.14$ and $1.64$ points below \gma{}, respectively; neither difference is CI-separated. At 8 steps, it gives up the routed arms' $2.14$-point advantage and scores $1.48$ points below \gma{} (\ns). The difference from each routed arm is $-3.62$ points (CIs $[-6.35,-0.88]$ and $[-6.24,-1.00]$). Tool-calling remains $16.37$ points above \gma{} at 24 steps (CI $[13.93,18.81]$), with gains of $17.66$ and $17.18$ points at 8 and 48 steps. Relative to \base{}, chain-of-thought accuracy changes by $+0.48$ points (\ns) and direct-answer accuracy by $+12.47$ points (\cisep). Multi-turn accuracy is $5.50$ points above \base{} and $2.0$ points below \gma{} (CI $[-7.88,+3.88]$), neither CI-separated.

On the first seed, no paired \gmd{} comparison is CI-separated below \gma{}, while tool-calling is CI-separated above it. This does not establish equivalence: the math point estimates remain below \gma{} at every step count. The math objective directly imitates \gma{}, which is the reference target for that route. The remaining $1.5$--$2.1$-point deficit may reflect incomplete convergence because math receives one third of the groups, or interference from tool-calling reward through shared parameters. The current experiment does not distinguish these explanations. An isolated math run would test whether cross-domain interference causes the gap.

\textbf{The tool-calling and preservation results replicate; math does not.} We repeated \gmd{} with identical prompt slices and changed only the run seed, then re-scored \base{}, \gma{}, and both seeds in one sweep. Before the new cells were analysed, the sweep reproduced all 12 prior \base{} and seed-one cells exactly in accuracy and item count. Seed two reproduces the tool-calling gain: $+18.23$ points over \gma{} at 24 steps (CI $[15.69,20.76]$), with gains of $19.27$ and $18.87$ points at 8 and 48 steps. The two preservation evaluations agree within $0.19$ points, and workflow and multi-turn performance hold on both seeds. Math does not replicate. At 8 steps, seed two is $3.95$ points below \gma{} (CI $[-6.87,-1.03]$), compared with $1.48$ points below for seed one (\ns). The math seed spread is $2.14$--$2.47$ points at 8 and 24 steps (Table~\ref{tab:seeds}). The seed-one 8-step margin is smaller than this spread, so a single seed could not resolve the no-loss criterion. The tool-calling gain reproduces across these two seeds; the math result is seed-dependent.

\begin{table}[ht]
\centering
\small
\caption{Replication of \gmd{}. The prompt slices and recipe are fixed, and
only the run seed changes. All arms are re-scored in one sweep, which first
reproduces every prior cell exactly. Tool-calling and preservation replicate;
math does not.}
\label{tab:seeds}
\begin{tabular}{lllr}
\toprule
cell & seed 1 & seed 2 & gap \\
\midrule
tool-calling @24 vs \gma & $+16.37$ \cisep & $+18.23$ \cisep & $1.86$ \\
math @8 vs \gma          & $-1.48$ \ns    & $-3.95$ \cisep{} below & $2.47$ \\
math @24 vs \gma         & $-2.14$ \ns    & $+0.00$ \ns & $2.14$ \\
math @48 vs \gma         & $-1.64$ \ns    & $-1.48$ \ns & $0.16$ \\
preserve CoT vs \base    & $+0.48$ \ns    & $+0.29$ \ns & $0.19$ \\
preserve answer vs \base & $+12.47$ \cisep & $+12.33$ \cisep & $0.14$ \\
\bottomrule
\end{tabular}
\end{table}

\section{Analysis}
\label{sec:analysis}
A fixed loss coefficient does not maintain a fixed anchor strength in this run. The remaining math differences vary with denoising steps but are smaller than the observed seed variation.

\textbf{The coefficient required for a fixed anchor ratio varies by $216\times$.} The gradient-norm ratio is fixed at $0.1000$ for all 240 steps, but the raw anchor-to-primary norm ratio ranges from $0.22$ to $47.8$ (median $1.64$). A coefficient selected at one step can therefore differ by up to two orders of magnitude from the value required at another. The inter-objective cosine ranges from $-0.997$ to $+0.994$, is negative on $38\%$ of steps, and tracks the batch route composition. This is a within-run measurement and is not affected by the between-arm confound in \S\ref{sec:method}.

\textbf{The apparent step dependence is not resolved.} \gmc{} adds the anchor but also removes the fallback route, which covers $1.8\%$ of math groups in \gmb{}. The design cannot isolate these changes. Relative to \gmb{}, \gmc{} improves math by $0.33$ points at 24 denoising steps and reduces it by $0.82$ points at 48 steps. Both differences are inside the $2.14$--$2.47$-point seed spread measured in \S\ref{sec:results}. We therefore report step dependence as a hypothesis rather than an established property of anchoring.

Concurrent work suggests one mechanism: KL to an imperfect reference can also preserve the reference model's errors~\citep{cokl2026}, with larger costs at high step counts where refinement proceeds further. A second explanation is conflict between reward and the anchor. \gmd{} provides a partial comparison because it uses stronger imitation of the same reference and no reward on math canvases, although tool-calling reward still updates shared parameters. At 48 steps, \gmd{} is $1.64$ points below \gma{}, compared with $3.45$ points for \gmc{}. This direction is consistent with reward--anchor conflict, but the $1.81$-point difference remains within the observed seed spread. A correctness-conditioned anchor would provide a more direct test.

\section{Limitations}
\label{sec:limitations}
We test one dLLM, and only the composed recipe has a second seed. The anchor-scale effects in \S\ref{sec:analysis} are about one point, below the measured seed variation, and are confounded by removal of the fallback route.

Each arm uses a different, disjoint slice of the math and workflow prompts. Differences between arms on these domains therefore combine the training recipe with prompt-slice variation. The replication fixes the slices, so its math spread measures seed variation alone. Tool-calling does not have this issue because every arm uses the same 350 training prompts.

Workflow has no binary verifier and is reported as a continuous macro score, so we make no paired significance claim. Tool-calling training is single-turn; the multi-turn set measures transfer from that setting. A confidence interval that includes zero does not establish equivalence. The composed arm's math point estimates remain below \gma{} at every step count. We report non-separation, not equivalence.

\section{Conclusion}
We present a likelihood-free dLLM post-training pipeline that combines multi-teacher on-policy distillation, outcome-routed REINFORCE, and a gradient-norm-scaled KL anchor. Routing improves tool-calling by $18.31$ points over the distilled checkpoint when the teacher supplies a successful target for only $0.8\%$ of the groups that the student fails completely. The same arm lowers chain-of-thought accuracy by $0.71$ points while raising direct-answer accuracy on the same items by $13.21$ points. The anchored arm restores chain-of-thought accuracy, with a $0.16$-point tool-calling difference whose confidence interval includes zero, although the comparison also removes the fallback distillation route.

A domain-conditional recipe retains tool-calling gains of $16.37$ and $18.23$ points across two seeds and reproduces both preservation results. Its math results remain seed-dependent. Across the three domains, teacher coverage of the student's failures tracks whether routing helps. This suggests routing between distillation and verifier RL using coverage measured on the student's own rollouts. The anchor and math results require more seeds and an isolated anchor comparison.

\bibliographystyle{plainnat}
{\small\bibliography{refs}}

\newpage

\end{document}